# Cube-Cut: Vertebral Body Segmentation in MRI-Data through Cubic-Shaped Divergences

Robert Schwarzenberg [a], Bernd Freisleben [a], Christopher Nimsky [b], Jan Egger [a,b]

[a] *Department of Mathematics and Computer Science, University of Marburg, Marburg, Germany*
[b] *Department of Neurosurgery, University Hospital of Marburg, Marburg, Germany*

Correspondence and requests for materials should be addressed to J.E.

egger@med.uni-marburg.de

# Abstract

In this article, we present a graph-based method using a cubic template for volumetric segmentation of vertebrae in magnetic resonance imaging (MRI) acquisitions. The user can define the degree of deviation from a regular cube via a smoothness value $\Delta$. The Cube-Cut algorithm generates a directed graph with two terminal nodes (s-t-network), where the nodes of the graph correspond to a cubic-shaped subset of the image's voxels. The weightings of the graph's terminal edges, which connect every node with a virtual source s or a virtual sink t, represent the affinity of a voxel to the vertebra (source) and to the background (sink). Furthermore, a set of infinite weighted and non-terminal edges implements the smoothness term. After graph construction, a minimal s-t-cut is calculated within polynomial computation time, which splits the nodes into two disjoint units. Subsequently, the segmentation result is determined out of the source-set. A quantitative evaluation of a C++ implementation of the algorithm resulted in an average Dice Similarity Coefficient (DSC) of 81.33% and a running time of less than a minute.

# Introduction

Lumbar stenosis (LS), a narrowing of any part of the lumbar spinal canal with encroachment on the neural structures by surrounding bone and soft tissue [1, 2] is the most frequent reason for surgery in patients over 65 years of age [1]. While MR imaging (MRI) is considered particularly purposive for the visualization of the soft tissue, X-ray computer tomography (CT) is seen as the method of choice for preoperatively evaluating bone anatomy [3]. CT, however, exposes the patient to carcinogenic radiation while the magnetic field in MR imaging is harmless.

Sometimes, degenerative spondylolisthesis, an asymptomatic slipping forward of one lumbar vertebra on another one with an intact neural arch, can be linked to LS [1] (Figure 1). Similar to LS, degenerative spondylolisthesis primarily occurs in elderly patients, and a combination of MRI and CT is also applied for preoperative evaluations in this case. A shift towards a more frequent application of MRI, even for morphological evaluations of the bone structure, would result in less radiation exposure [3], which is also what motivates this work

Several features of the spinal anatomy can be distinguished by their different grey values in an MR image. In most T1- and T2-weighted image slices, normal adult vertebral body bone marrow can be differentiated from the outer boundaries of the vertebral body by a homogeneously lighter grey value [4]. This is because the outer, compact cortical bone, which coats the vertebral body, results in a much darker color/lower grey value than the cancellous, spongy inner part. Thus, the grey-value difference between a voxel in the vertebral body and a voxel on the outer boundaries (e.g. cortical bone) is higher than the difference between two voxels inside the vertebral body.

This, however, does not apply to slices that depict the pedicles. Figure 2 shows that the pedicles of the vertebral arch are not considered part of the vertebral body. Nevertheless, since they are connected to the vertebral body, they belong to its outer boundaries. However, unlike the cortical bone, they define a weak, homogeneous object-background transition region. Furthermore, in Figure 3, instead of the cortical bone, the cerebrospinal fluid (CSF, surrounding the red arrow), which causes the high grey value of the spinal canal in T2-weighted images [4], defines parts of the outer boundary of the vertebral body. This is due to noise and signal distortion, resulting in an overlapping. Due to signal distortion and noise, as well as anatomical structures like the pedicles and occasional voxel outliers, the vertebral body cannot be defined by sharp boundaries in all MR image slices - a challenge every segmentation algorithm has to address.

Several approaches for vertebra segmentation have been proposed in the literature [6-21]. Some of them [6-11] belong to the 2D approaches and others [12-21] belong to the 3D approaches. As we present a novel volumetric approach in this contribution, the following state-of-the-art paragraphs introduce the 3D approaches in greater detail. [12-18] all use some kind of shape constraints and the shapes in [12, 15, 18] rely on training data. In contrast, [19] is a free-form segmentation approach which uses balloon forces. At the end of this background section, we discuss a training-based model which detects and labels intervertebral disks in MR images [21].

Klinder et al. [12] use articulated shape models for spine segmentation. Their approach considers not only individual objects (vertebrae) one at a time, but also object constellations from a more global perspective. A constellation is presented as a consecution of local vertebra coordinate systems, whereas the individual morphology of a vertebra is encoded into a triangulated surface model. Performing non-rigid deformations, the processing of an individual object happens simultaneously to the processing of all other objects, allowing different deformation processes to interact. Klinder et al. explain that in the course of this, the model is attracted to image features but that the attraction is also constrained by a former learned shape. For their method, the authors report a segmentation accuracy of 1.0 mm in average for ten thoracic CT images.

Hoad and Martel [13] describe a three-step algorithm, which segments bone from soft tissue in MR images of the spine. In the first step, the vertebral bodies are segmented, in the second step the posterior structures are segmented, and in the third step, manual corrections are made. The authors explain that in the three different stages, they combine thresholded region growing with morphological filtering and masking using set shapes. For evaluation, they registered the segmented data to a physical model of a spine which they obtained using CT scans. Hoad and Martel report that their method produces segmentation results equally suitable for registration as the *gold standard* CT data and they regard their algorithm as robust. Furthermore, they point out that different threshold levels, within visually acceptable intervals, had very little effect on the registration results. The authors conclude that in general, the accuracy of the registration relies on the similarity between actual and automatically generated surfaces as well as the precision of the digitized points used for the registration.

Štern et al. [14] make use of superquadrics to deterministically model volumetric shapes of vertebral bodies which they then align with vertebrae in 3D CT scans and MR images, for segmentation. All in all, they introduce 29 parameters to the superquadric function to obtain a vertebral-shaped geometrical model. In case the user wants to incorporate certain pathological deformations, further parameters have to be introduced. The parameters are then automatically optimized in order to achieve the most accurate alignment of the model with the vertebral body in the CT or MRI data. The optimization is driven by a combination of intensity gradient information with image intensity appearance of the bone structures and surrounding soft tissues. The method is initialized with a single point inside the vertebral body and was tested on 75 vertebrae from CT scans and 75 vertebrae in MR images. Štern et al. performed 100 segmentation experiments per vertebra by randomly displacing the initial 3D model from the ground truth pose and considered the subsequent segmentation successful if the mean radial Euclidian distance of the final 3D model from the ground truth points was less than 3 mm. For their experiments, they report an overall mean radial Euclidian distance (±standard deviation) between the final 3D models and the ground truth points of 1.17±0.33 mm for CT images (success rate 94.5%) and 1.85±0.47 mm for MR images (success rate 88.6%).

Aslan et al. [15] describe a graph-based method for the volumetric segmentation of vertebral bodies which incorporates shape priors. The authors obtain the required shape information from a training set of manually segmented vertebral bodies in CT data: After aligning the manual segmentation results, they determine an object region that describes the cross section of all vertebral bodies, a variability volume, consisting of the remaining target-structure voxels, and a background region. To detect shape variations in the variability zone, Aslan et al. apply a distance probabilistic model. Then, they construct an undirected, weighted graph, implementing the 3D-shape prior through the edges' capacities. In a final step, a minimum cost cut is performed, partitioning the image's voxel set into two disjoint units, namely the target structure and the background.

Weese et al. [16] present shape constrained deformable models for 3D medical image segmentation, which they apply to vertebra CT acquisitions. Their hybrid approach combines the advantages of an active shape model and an elastically deformable surface model. The latter one is implemented as a surface mesh, whereby its flexibility is constrained by the shape model, which also ensures an optimal distribution of mesh vertices. In order to increase their approach's robustness against false object boundaries, Weese et al. attract the deformable model to locally detected surfaces, using an external energy. For validation, they compared the semi-automatic segmentation results of their algorithm to manually segmented vertebrae. In case of a proper manual placement of the mean vertebra model, Weese et al. report a mean segmentation error of 0.93 mm with deviations around 4.5-7 mm in problematic areas.

Yao et al. [17] describe a method for the automatic segmentation and partitioning of the spinal column. Their approach starts with a simple thresholding to mask out bone voxels and subsequently, it applies blob extraction, identifying the largest connected blob as the initial spine segmentation. Yao et al. explain that afterwards, a hybrid method based on the watershed algorithm and directed graph search is employed to obtain the spinal canal. They then use the spinal canal to position a vertebra model which consists of four parts, namely the vertebral body, spinous process, and left and right transverse processes. In the next step, the initial model is deformed in a way such that a maximum model-to-image match is achieved. In the last step they generate curved planar reformations (CPRs) in sagittal and coronal directions as well as they analyze aggregated intensity profiles along the spinal cord in order to partition the spinal column into the different vertebrae. For evaluation, the approach was tested on 71 CT scans and the authors state that the algorithm successfully extracted and partitioned 69 spinal columns, with only 2 cases that had one missed partition at the T1-T2 level.

Ghebreab and Smeulders [18] present an integral deformable spine model for three-dimensional segmentation of spinal images. They explain that their approach learns the representation of vertebrae in CT scans from multiple continuous features registered along vertebra boundaries in a given training set. Statistics are encoded into a necklace model, which is coupled by string models that provide detailed information on morphological variations in the appearance of spinal structures from multiple continuous features registered in the training set. Ghebreab and Smeulders further state that on the necklace model, landmarks are differentiated on their free dimensions and that, in order to reduce complexity, the landmarks are used within a priority segmentation scheme. For segmentation of new image data, the necklace and the string models are employed to detect vertebral structures interactively by means of elastic deformations. Ghebreab and Smeulder remark that this bears an analogy to a marionette with strings constraining the deformations in a way such that only movements within feasible solutions are allowed.

Zukić et al. [19], [20] present a fast and semi-automatic approach for spine segmentation in routine clinical MR images. A single vertebra is segmented based on multiple-feature boundary classification and mesh inflation, and it starts with a simple point-in-vertebra initialization. To prevent self-intersections, the inflation retains a star-shaped geometry and the smoothness is controlled via a constrained subdivision hierarchy. The main spine direction is deduced by analyzing the shape of the first vertebra and the locations of neighboring vertebral bodies are estimated for further segmentation. Against manual reference segmentations, the average Dice Similarity Coefficient (DSC) [22, 23] was 78% and a detection rate of 93%. The approach was tested on eleven routine lumbar datasets with 92 segmented vertebrae.

Kelm et al. [21] use iterated marginal space learning (MSL) to detect and label intervertebral disks in MR images. Furthermore, they claim that since their approach is learning-based, it can be applied to CT scans, as well. In a first step (after roughly locating the spine), their method uses an iterative extension of the MSL method to determine candidate regions including the potential targets' positions, orientations, and scales. In a second step, Kelm et al. use a global probabilistic spine model to detect the most probable candidates among them. They report that experimental validations of their method revealed 98.6% sensitivity, 7.3% false positive detections, an average position error of 2.4 mm, an angular error of 3.9°, and an overall processing time of 11.5 seconds.

Our approach to solve the problem of three-dimensional vertebral body segmentation is a non-trivial enhancement of the previously introduced two-dimensional graph-based segmentation strategy *Square-Cut* [11], which uses a rectangle template to segment vertebral bodies on single MRI-slices. Consequently, we now use a cubic-shaped distribution of the graph's nodes in the three-dimensional case. Moreover, we developed, implemented and evaluated far more complex three-dimensional neighborhood relations which can be easily altered as they are implemented as a function of a user-defined smoothness-term.

The rest of this contribution is organized as follows: *Section 2* presents the methods behind the introduced algorithm, *Section 3* presents the results of our experiments and *Section 4* concludes the paper and outlines areas for future work.

## Methods

The new vertebral body segmentation algorithm presented here will be referred to as *Cube-Cut*. Cube-Cut extends a two-dimensional approach, previously introduced by Egger et al. [11] to a third dimension (Note: initial results of *Cube-Cut* have been presented on a workshop [24] and a German conference [25]). This extension allows the volumetric segmentation of a vertebral body with only one click, instead of just a two-dimensional segmentation on a single slice. The introductory paragraphs of this section first give a conceptual overview of the basic features and the behavior of Cube-Cut. This conceptual overview serves as a frame of reference for the more detailed discussion of the actual implementation that follows at a later stage and which introduces the reader to the concepts of a cubic-shaped graph and the related smoothness-constraint. To keep it consistent, the notation for the graph construction follows the notations of previous graph-based publications [26-35], where possible.

**2.1 Conceptual Overview**

2.1.1 Labeling

Given a volumetric MR image $P$, Cube-Cut first selects a subset $P' \subseteq P$ of the image's voxels and in a last step it tags each voxel $p \in P'$ with either one of the labels $L_s$ or $L_t$ [36]:

$$t : P' \rightarrow \{L_s, L_t\} \tag{1}$$

2.1.2 Penalties

The labeling of a voxel $p \in P'$ involves two penalties [36]:

- $D_p(t(p)) \in R_{\geq 0}$ is the penalty for assigning the label $t(p)$ to $p$ and
- $V_{p,p'}(t(p),t(p')) \in R_{\geq 0}$ is the penalty for assigning $t(p)$ to $p$ when $t(p')$ is the label of the voxel $p' \in P'$.

$D$ describes a voxel's affinities to the labels $L_s$ and $L_t$. For example, the higher $D_p(t(p)=L_t)$, the more $p$ is affiliated with $L_s$. $V$ on the other hand reflects a voxel's affiliation to another voxel. In practice, $V_{p,p'}(t(p), t(p'))$ is greater than zero only if $t(p) \neq t(p')$. Thus, $V$ indirectly describes $p$'s affiliation with $p'$ by awarding a penalty for tagging the two voxels with different labels: The higher $V_{p,p'}$, the more p is affiliated with $p'$. Nevertheless, note that $V_{p,p'}(t(p), t(p'))=0$ for $t(p) \neq t(p')$ does not necessarily mean that the two voxels $p$ and $p'$ can be tagged differently without penalty costs. If, for instance, $V_{p,p''}(t(p), t(p''))>0$, for $t(p) \neq t(p'')$ and $p'' \in P'$ and if $V_{p'',p'}(t(p''), t(p')) > 0$ for $t(p'') \neq t(p')$, then the penalty cost for assigning different labels to $p$ and $p'$ is at least $min\{V_{p,p''}, V_{p'',p'}\}$.

2.1.3 Return Value

Cube-Cut tags the voxels in a way such that the overall penalty cost is minimized. The overall cost is described by (2). Cube-Cut thus returns the argument $L$ which minimizes

$$E(L) = \sum_{\substack{p \in P' \\ t(p) \in L}} D_p(t(p)) + \sum_{\substack{p,p' \in P' \\ t(p),t(p') \in L}} V_{p,p'}(t(p),t(p')) \tag{2}$$

where $L=\{t(p)|p \in P'\}$ is a labeling of the subset $P'$ [36, 37]. However, until now, the features above have been discussed detached from the context of vertebral segmentation. The next section will make the connection.

2.1.4 Object and Background Separation

Cube-Cut selects $P'$ and implements $D$ and $V$ (on the basis of the predetermined subset $P'$) in a way such that the returned labeling is to be interpreted in the following manner (Figure 4):

- Cube-Cut assumes all voxels $p \in P'$ for which $t(p)=L_s$ inside the vertebral body and
- all voxels $p \in P'$ for which $t(p)=L_t$ can be assumed outside the vertebral body.

Hence, in a first step Cube-Cut selects a subset of voxels, and on the basis of this subset, implements two penalty functions which then determine a clustering of the subset into two disjoint units of voxels. One unit describes the vertebral body while the other describes the background (which may include other vertebrae). The following paragraphs will describe the implementation of the algorithm.

**2.2 Implementation**

2.2.1 Voxel Subset

The voxels $p \in P'$ are distributed along $n$ rays that expand from a user-defined seed point in the MR image. Each ray consists of $k$ equidistantly spread voxels, where for all rays, the first voxel is always the user-defined seed point, so that $|P'| = n * (k - 1) + 1$. As the seed point, the number and the length of the rays as well as the number of voxels per ray can be determined by the user. It is assumed that each ray exceeds the vertebral body. In the following, let $p_{i_r} \in P'$ denotes a voxel on ray $r$, where $1 \leq r \leq n$ and where the voxel $p_{i_r}$ is closer to the seed point ($p_1$ or $p_{1_r}$) than $p_{j_r}$, if $1 \leq i < j \leq k$ (Note: If only one ray is being discussed, the indexing $r$ might be omitted. Furthermore, from now on, $k$ will always denote the number of voxels per ray and $n$ the number of rays).

2.2.2 Cubic Distribution

The rays expand in a way such that all voxels of the same layer form a cube shape, so that if $i = j \neq 1$ and $m \neq n$, then the voxels $p_{i_m}$ and $p_{j_n}$ lie on the surface of one cube, which has the user-defined seed point as its center. Since there are $k$ voxels on each ray, there are $k-1$ different sized cubes for which $p_1$ is the center (Figure 5). On a cube's face, the voxels are distributed equidistantly and the volumes of the cubes increase evenly. However, note that due to the theorem of intersecting lines, the distance between two voxels on a cube's face is less than the distance of the corresponding voxels on a bigger cube.

2.2.3 Implementation of Penalties and Labeling

Cube-Cut generates a network $N = ((G = (V (G), E(G))), c, s, t)$, where $G$ is a directed, two-terminal graph and $|V (G)| = |P'| + 2$. Each vertex $v \in V(G) \backslash \{s,t\}$ corresponds to exactly one voxel $p \in P'$ and no two vertices correspond to the same voxel. In the following, $v_p$ will denote a mapping of the vertex $v \in V(G) \backslash \{s,t\}$ onto its corresponding voxel $p \in P'$ and $p_v$ will describe the reverse mapping. The source $s$ and the sink $t$ have no counterparts in $P'$ and thus they are referred to as *virtual* nodes. In $E(G)$, there exist two types of edges [36], [38] (Figure 6):

- *i-links* (inter-links) connect vertices $v \in V(G) \backslash \{s,t\}$ with each other. The *i-links* are further subdivided into *z-edges* and *xy-edges*, where *z-edges* connect vertices corresponding to neighboring voxels of the same ray (e.g. ($v_{i_n}, v_{(i+1)_n}$)), while *xy-edges* connect vertices corresponding to voxels of different rays (e.g. ($v_{i_n}, v_{j_m}$)).

- *o-links* (outward-links) connect all vertices $v \in V(G)\setminus\{s,t\}$ with the source *s* (*s-links*) and the sink *t* (*t-links*). Hence, there are two *o-links* for each vertex.

The capacities of the *i-* and *o-links* reflect the penalty functions *D* and *V* in the following manner (Figure 7):

$$\forall v \in V(G) \setminus \{s,t\} : c(s,v) = D_p((t(v_p) = L_t)),$$
$$\forall v \in V(G) \setminus \{s,t\} : c(v,t) = D_p((t(v_p) = L_s)), \quad (3)$$
$$\forall (v,v') \in E(G) : c(v,v') = V_{p,p'}(t(v_p), t(v'_p)).$$

Note that the skew symmetry constraint does not have an effect since by convention *c(v, v') = 0* is assumed, if *(v, v')* ∉ *E (G)*. After the graph has been set up, Cube-Cut determines a minimal s-t-cut (S, T) by deploying the Boykov-Kolmogorov algorithm [36] (http://vision.csd.uwo.ca/code/, accessed: March 2014) and then it labels P' as follows:

$$\forall v_p \in P' : t(v_p) \begin{cases} L_s & \text{if } v \in S; \\ L_t & \text{else.} \end{cases} \quad (4)$$

Since by definition, the capacity of a minimal s-t-cut is minimal among all possible s-t-cuts, the labeling above minimizes (2).

2.2.4 Z-Edges: Onetime Cut per Ray

Since each ray intersects with the outer boundaries of the vertebral body only once, a set of *z-edges* is introduced that ensures that each ray is exactly cut one time by a minimal s-t-cut [38, 39]:

$$A_z = \{(v_{i_r}, v_{(i-1)_r}) | 1 < i \leq k \land 1 \leq r \leq n\}, \quad (5)$$

where *n* is the user-defined number of rays and *k* the total number of voxels per ray, again (note: in what follows, $v_{i_r}$ will denote the corresponding vertex of the $i^{th}$ voxel on ray *r*. Furthermore, if only one ray is being discussed, the indexing *r* might be omitted). The set of *z-edges* connects each vertex $v_i$ with its predecessor $v_{(i-1)}$ on the same ray (Figure 8). The capacities of all *z-edges* are initialized to ∞. Therefore, it costs ∞ each time a *z-edge* is cut.

By making sure that the seed point is in *S* and that the last voxel on each ray is in *T* (see next section), a minimal s-t-cut (*S, T*) has to cut each ray at least once. Yet, it does not cut any ray more than once because that would cost at least 2·∞. This is why a ray is cut exactly one time. The next section explains how Cube-Cut encourages this cut to happen close in front of the vertebral body's outer boundaries.

2.2.5 O-links: Marking the Outer Boundaries

A voxel $p_{i_r}$ is characterized by $(x_{i_r}, y_{i_r}, z_{i_r}, g_{i_r})$ where $x_{i_r}, y_{i_r}, z_{i_r} \in \mathbb{N}_0$ denote the voxel's position in the image and $g_{i_r} \in \mathbb{R}_{\geq 0}$ denotes its grey value (note: simplified, voxel coordinates assumed). Cube-Cut investigates a small cube $((x_1, y_1, z_1), (x_2, y_2, z_2))$ around the user-defined seed point (inside the vertebra) and determines its interval of grey values $I = [min(GV), max(GV)]$, where

$$GV = \{\pi_4(x, y, z, g) \mid x_1 \leq x \leq x_2, y_1 \leq y \leq y_2, z_1 \leq z \leq z_2\} \qquad (6)$$

is the multi-set of all grey values within the cube and $\pi_i(\cdot)$ is a projection onto the $i^{th}$ element of a tuple. Furthermore, Cube-Cut also iterates over the cube to determine an average grey value $g_{avg}$ by

$$g_{avg} = \frac{1}{|GV|} \cdot \int_{x_1}^{x_2}\int_{y_1}^{y_2}\int_{z_1}^{z_2} \pi_4(x, y, z, g)\, dx\, dy\, dz. \qquad (7)$$

In the course of weighting the *o-links*, the interval $I$ and the average grey value $g_{avg}$ are used as frames of reference.

The following Pseudo-Code depicts the fundamental principle of how Cube-Cut assigns capacities to the *o-links* (note: Each ray $r$ consists of $k$ voxels):

0      $c(s, v_1) \leftarrow \infty$

1      $c(v_1, t) \leftarrow 0$

2      *assign(ray r)*

3      $\forall p_{i_r} = (x_i, y_i, z_i, g_i) \in r \setminus \{p_{1_r}\}$

4         if $i == k$

5             $c(s, v_{i_r}) \leftarrow 0$

6             $c(v_{i_r}, t) \leftarrow \infty$

7         *else if* $g_i \in I$ *or* $abs(g_{avg} - g_i) \leq abs(g_{avg} - g_{i-1})$

8             $c(s, v_{i_r}) \leftarrow abs(abs(g_{avg} - g_i) - abs(g_{avg} - g_{i-1}))$

9             $c(v_{i_r}, t) \leftarrow 0$

10        *else*

11            $c(s, v_{i_r}) \leftarrow 0$

$$c(v_{i_r}, t) \leftarrow abs(abs(g_{avg} - g_i) - abs(g_{avg} - g_{i-1}))$$

The $\infty$-weighting in line 0 ensures that the seed point is tagged with $L_s$. The premise on which this is based is that the user defines the seed point within the vertebral body (in the center).

Furthermore, it is assumed that the last voxel on each ray ($p_{k_r}$) lies outside the vertebra (since the user is supposed to define a ray length that exceeds the vertebral body). To ensure that the last voxels are tagged with $L_t$, the *t-links* ($v_{k_r}$, t) are also $\infty$-weighted while for all rays $c(s, v_{k_r})$ is consequently initialized to zero (line 4 - 6).

The capacities of all of the other, intermediate *o-links* reflect the value difference between a voxel and its predecessor on the ray (lines 8,9,11 & 12). This is in order to "mark" the outer boundaries.

As already mentioned above, the rays expand from the user-defined seed point in the center of the vertebral body and they eventually intersect with the outer boundaries. Ignoring occasional outliers and homogeneous object/background transition regions for now, the inner vertebral body is characterized by a homogeneous set of voxel grey values, which are all higher or lower than the grey values that make up the outer boundaries (e.g. cortical bone, spinal canal, compare *Introduction*). Thus, on each ray, the difference in value between the last voxel in the vertebral body and the first voxel on the outer boundaries can be assumed high.

Taking the condition in line 7 into account, the outer boundaries therefore implement high *t-link* capacities (line 12). Note that this makes a cut right in front of the corresponding vertices very probable. The next sections explain how peculiarities and anomalies in vertebral MRI data sometimes prevent a cut from happening right in front of the outer boundaries and how Cube-Cut addresses these adverse effects.

2.2.6 Adverse Effects on the Segmentation Result

A cut right in front of the outer boundaries is a cut that separates the last vertex that corresponds to a voxel which is still located inside the vertebral body from the subsequent ones on the same ray. If, for each ray, the cut takes place right in front of the outer boundaries of the vertebral body, then Cube-Cut returns a satisfactory segmentation result.

Let $v_{i_r}$ be the first vertex on a ray $r$ that corresponds to a voxel on the outer boundaries/background. If a minimal s-t-cut (*S, T*) cuts the ray right in front of $v_{i_r}$, so that $v_{(i-1)_r} \in S$ and $v_{i_r} \in T$, then

$$\sum_{j=i}^{k-1} c(s, v_{j_r}) < \sum_{j=i}^{k-1} c(v_{j_r}, t) \qquad (8)$$

and

$$\forall h < i : h > 1 \Rightarrow \sum_{j=h}^{i-1} c(v_{j_r}, t) \leq \sum_{j=h}^{i-1} c(s, v_{j_r}) \qquad (9)$$

For most rays, the two (minimum) conditions hold true and thus, the cut takes place right in front of the outer boundaries. Equation (8) usually holds true because the value difference between $p_{i_r}$ and $p_{(i-1)_r}$ is greater than the sum of the subsequent *s-weights* since behind the outer boundaries, the rays mostly penetrate homogeneous areas dissimilar from the vertebral body (compare lines 7 & 8 of the pseudo code). Equation (9) holds true for most rays because of the homogeneity of voxel grey values in the vertebral body and their similarity to the close environment of the seed point (compare lines 7, 8 & 12 of the pseudo code).

Nevertheless, there are exceptions. Figure 9 depicts such exceptions. (a) clearly shows a 2-dimensional view of a segmentation result that overruns the vertebral body in the upper part. For the corresponding rays, equation (3) does not hold true. The similarity between the vertebral and the intervertebral voxels, in terms of their grey values, can easily be recognized. Furthermore, there are minor variations of grey values in the intervertebral disc.

As a consequence, the condition in line 7 of the pseudo code holds true for a sufficient number of background voxels on each of the affected rays, which is why condition (8) is not satisfied. Thus, the overrun occurs. Observe that the same applies to homogeneous object/background transition regions. Among others, Cube-Cut tackles this problem by introducing a coefficient *w*, which loads the *s-weights* according to their distance from the seed point (see next section). Line 8 of the pseudo code is extended to:

$$c(s, v_{i_r}) \leftarrow w(i,k) \cdot abs(abs(g_{avg} - g_i) - abs(g_{avg} - g_{i-1}))$$

Another phenomenon that negatively affects the segmentation result is outliers. Outliers share all relevant properties (grey values) that distinguish the vertebra's boundaries except that they are part of the inner vertebral body.

To be specific, an outlier causes the violation of equation (9). On the corresponding ray, the cut then happens too close to the seed point (Figure 9 (b)). Cube-Cut decreases the possible adverse effects due to outliers by imposing a smoothness constraint on the segmentation result. In addition, the smoothness constraint also addresses the problem of a violation of equation (8), as discussed above. The next two sections present Cube-Cut's problem-solving approaches in detail. The first matter to be addressed will be the loading of the *s-capacities* and then the smoothness constraint will be discussed.

2.2.7 Loading the s-Capacities

The coefficient $w(\cdot)$ loads an *s-capacity* according to the corresponding voxel's ($p_{i_r}$) position on the ray (Figure 10). For a ray *r*, consisting of *k* voxels, it is defined as $w(i,k) = mi+b$, where $k \geq i \in \mathbb{N}_{>0}$ and $m = -\dfrac{1}{k-1}$ and $b = 1-m$.

Observe that since the voxels are distributed uniformly on each ray, $w(i, k) = 1$ for the seed point ($p_{i=1}$), $w(\cdot, k) = 0.5$ for a voxel that is half way on a ray (Figure 11) and $w(\cdot, k) = 0$ for the last voxel on each ray. A voxel far away from the seed point is more likely to be outside the vertebral body.

Cube-Cut takes this into account by decreasing its *s-capacity* accordingly, thereby reducing the risk of a cut being located behind the outer boundaries of the vertebral body because

$$\sum_{j=i}^{k-1} c(s, v_{j_r}) > \sum_{j=i}^{k-1} w(j,k) \cdot c(s, v_{j_r}) \qquad (10)$$

As already mentioned above, the coefficient is not the only measure Cube-Cut takes in order to counteract a violation of condition (8): The smoothness constraint, which also addresses a violation of condition (9), will be the subject matter in the next section.

2.2.8 XY-edges: Imposing a Smoothness Constraint

The smoothness constraint is based on the optimal surface segmentation algorithm developed by Li et al. [38]. It is useful to first discuss it conceptually, slightly detached from the context of vertebral segmentation. A single, feasible surface in a volumetric Image $I = (X, Y, Z)$, where $X, Y, Z \subset N_0$, can be characterized by a bijection $S : XY \rightarrow Z$, where $XY \subseteq X \times Y$ is a cohesive area. Li et al. refer to a surface as feasible if two smoothness constraints are satisfied:

$$\forall (x,y), (x+1, y) \in XY : | S(x,y) - S(x+1, y) | \leq \Delta_x \qquad (11)$$

and

$$\forall (x,y), (x, y+1) \in XY : | S(x,y) - S(x, y+1) | \leq \Delta_y \qquad (12)$$

$\Delta_x$ and $\Delta_y$ constrain the degree to which the surface "moves" upwards or downwards in x- or y-direction within an interval of one: $S(x, y)$ and $S(x+1, y)$ as well as $S(x, y)$ and $S(x, y+1)$ are neighboring x- and y-positions. Thus, two neighbors on a feasible surface cannot be arbitrarily distant from each other. Hereby, the smoothness constraints assure what Li et al. refer to as "surface connectivity". Observe that for a plane $\Delta_x = \Delta_y = 0$.

Now consider a number of equidistant rays that consist of the same number of uniformly spread voxels and which all extend parallel to the *z-axis*. The voxels that make up a ray do not necessarily have to lie on neighboring positions in the image. Furthermore, for convenience, assume that *I* is a binary image with only two possible values for each voxel: "colored" *xor* "white". In addition, let all rays intersect with a colored surface *S(XY)* in *I*, which means that each ray extends from *XY* and shares exactly one colored voxel with the surface.

In this context, in which only a subset of the image's voxels is observed, the smoothness constraint has to be defined via the neighborhood relations of the rays. Figure 12 shows four neighboring rays in *x-direction* (same *y-value* for each voxel), which extend parallel to the *z-axis*, as described above. Here, a smoothness parameter $\Delta_x = 1$ means that for a "colored" voxel that is considered part of the surface, all voxels on adjacent rays that are also classed with the surface voxels must lie on the same "z-layer" or the next upper or lower one. An outlier in this context is a colored voxel that exceeds the prescribed maximum distance.

Cube-Cut allows the user to impose a smoothness constraint on the segmentation result. It interprets each of the six sides of a vertebral body's outer boundaries (from a sagittal view: front, back, top, bottom, right, and left) as a feasible surface. Furthermore, it takes into account that the six surfaces are anatomically connected, which is why the neighborhood relations overlap at the "edges" of the boundaries.

Cube-Cut implements the smoothness constraint $\Delta \in N_0$ by introducing a set of infinity-weighted *xy-edges* [38]:

$$A_{xy} = \{(v_{i_r}, v_{(\max\{(i-\Delta),1\})_{r'}})\} \mid (r,r') \in N_4\} \tag{13}$$

$N_4$ denotes a 4-neighborhood and as already mentioned above, the neighborhood relations overlap at the "edges" of the cubic voxel subset that the algorithm observes.

The infinity-weighting of the *xy-edges* ensures that a minimal s-t-cut cuts the rays in a way such that the vertebra is segmented within the boundaries of the user-defined smoothness constraint Δ (Figure 13). Note that for a given Δ-value, an 8-neighborhood would increase the "stiffness" of the segmentation result.

For a smoothness constraint Δ=0, any minimal s-t-cut results in a regular, cubic segmentation result, whereas a Δ-value greater zero allows a corresponding deviation. Figure 14 shows the topology of the *xy-edges* for a Δ-value of zero and a Δ-value of one and Figure 15 shows corresponding segmentation results, illustrating the purpose of the cubic shaped voxel subset.

## Results

A C++ implementation of Cube-Cut was tested within the medical image processing platform MeVisLab 2.2.1 (www.mevislab.de). We used two T2-weighted, volumetric, pathological MR-images (512x512x16 and 512x512x10), both of which had an anisotropic voxel spacing in x- and y-directions of 0.63 millimeters and 4.4 millimeters along the z-axis. We obtained isotropic voxel sizing (2.01258 millimeters in all directions) by resampling the two images, using the MeVisLab Resample3D-module, which resulted in resolutions of 159x159x35 and 159x159x22 respectively. The first image contained a stenosis and a spondylolisthesis; the second image showed a slipped disc.

In order to place the seed-point roughly in the center of the vertebra, we scrolled through two-dimensional, sagittal image slices. After Cube-Cut terminated, the triangulation was visually evaluated. If overruns occurred, we decremented the smoothness-constraint and/or replaced the seed-point accordingly, e.g. we moved it in x-direction if the overrun occurred in y-direction. We learned that once parameter settings were found for one vertebra in a data set, these settings could also be successfully applied to most of the other vertebrae in the same image. Figure 16 shows volumetric and two-dimensional segmentation results.

The obtained segmentation results were then compared to ten segmentation results obtained in a slice-by-slice manner, performed by trained physicians. Table 1 presents the detailed results for all ten cases and in addition the summary of results, with min, max, mean $\mu$ and standard deviation $\sigma$. For visual inspection, Figure 17 shows a superimposition of a pure manually segmented and an automatically obtained (Cube-Cut) segmentation result. Furthermore, the manual slice-by-slice segmentations have been compared to segmentation results obtained using the GrowCut-algorithm [40] (Table 2). For testing GrowCut with our datasets we used an implementation that is freely available in the medical platform (3D) Slicer (www.slicer.org). For initialization of the GrowCut algorithm, strokes have been drawn inside and outside the vertebral body on a two-dimensional, sagittal image, a two-dimensional axial image and a two-dimensional coronal image (Figure 18), as it has been done in [41], [42] and [43]. Table 2 presents the direct comparison of the manual slice-by-slice and a GrowCut segmentation for the ten vertebrae from Table 1 with a mean DSC-value of 80.61%.

It was found that it takes a trained physician 10±6.65 minutes to manually segment a vertebra in a slice-by-slice manner. On a 2.1 GHz x64-based PC with 4 GB RAM running the Microsoft Windows 7 Home Premium (SP1) operating system, version 6.1.7601, the most expensive parameter settings took Cube-Cut less than a minute (graph-construction, mincut computation and triangulation) to terminate. The settings that resulted in a maximum DSC of over 86% only took 19 seconds to execute. Overall, we achieved a mean DSC-value of 81.33%.

**Conclusion**

A novel approach towards the volumetric segmentation of vertebral bodies was presented. Cube-Cut is a non-trivial, three-dimensional extension of the previously introduced two-dimensional segmentation strategy Square-Cut [11] and a proof of concept implementation of the optimal surface segmentation approach by Li et al. [38]. The introduced method is the first one using a 3D-graph that is based on a cubic-shaped subset of non-equidistant image voxels as well as a smoothness-constraint in order to segment volumetric, cubic-like target-structures. The possibility to approach a cubic template by changing the graph's topology as a function of the user-defined smoothness-term in real-time effectively allows overcoming homogeneous object-background transition regions. In summary, the research highlights are:

- development of a specific graph-based algorithm for vertebral body segmentation;

- algorithm bases on a cubic template which is a novelty in the segmentation domain;

- scale-invariant segmentation by an optimal mincut through cubic-shaped divergences;

- physicians performed slice-by-slice segmentations to obtain ground truth boundaries;

- segmentation quality of the algorithm has been evaluated via the Dice Coefficient.

The proposed method only requires a single user seed, while other approaches [13, 18] require multiple user-inputs to achieve comparable results. On the other hand, the easily alterable smoothness term seemingly provides more flexibility than the approach proposed by Štern et al. [14], since the authors state that they might have to introduce new parameters to their deterministic model when confronted with not yet considered pathologies or deformations.

In addition, contrary to most of the alternative approaches [12, 15, 18], our proposed method does not rely on training data and is thus not constrained to the variations, deformations and pathologies covered in the data set. This also means that Cube-Cut needs a far less expensive initialization phase. Furthermore, the graph-based approach presented by Aslan et al. [15] observes the whole set of the image's voxels, while Cube-Cut only requires a subset. Both algorithms compute the mincut in polynomial time and thus our approach outperforms the approach of Aslan et al. in terms of theoretical runtime.

Furthermore, whereas other approaches [12, 15, 16, 17, 18] have tested their algorithms only on CT datasets, we show that our approach is suitable for MR-image processing, and contrary to several other approaches [13, 14], we have tested our algorithm on pathological spine data, achieving a better mean DSC than Zukić et al. [19]. A shift towards a more frequent application of MRI in the preoperative evaluation of surgical patients would result in less radiation exposure compared to the more frequent application of CT.

To summarize, Cube-Cut generates a two-terminal s-t-network where the vertices correspond to a cubic shaped subset of the image's voxels. By only observing a subset of the image's voxels, the algorithm improves the theoretical runtime in comparison to other graph-based approaches that consult the whole voxel set. The capacities of the terminal edges reflect a voxel's affiliation with the object (vertebral body) and the background, while the topology of non-terminal ∞-weighted edges implements the smoothness-constraint. After network construction, a minimal s-t-cut, computed in polynomial time, determines the segmentation result. For the mincut computation, the Boykov-Kolmogorov algorithm is applied, since it was demonstrated [36], that despite a worst case complexity of $O(v(G)^2 e(G) /C_{min}/)$, where $v(G)$ is the number of vertices, $e(G)$ the number of edges and $/C_{min}/$ the capacity of a minimal cut, the algorithm is most effective for networks of low complexity, like the one Cube-Cut generates. In addition, we want to point out here that Boykov's graph cut model can support high dimensional data in its own settings.

For evaluation, Cube-Cut 3D-masks were compared to manual vertebral body segmentation results, obtained in a time-costly slice-by-slice manner, performed by trained physicians, achieving a promising mean DSC of 81.33 %, which is on par with the state of the art (comparable to [19]). The computation (graph construction, mincut computation and triangulation) that led to the maximum DSC value of 86.69 % terminated in 19 seconds on a customary PC. All other parameter settings took no more than a minute to execute. It was found that a slice-by-slice segmentation of a vertebra took trained physicians 10±6.65 minutes on average, and a subsequent conversion into a 3D-mask was also still needed. This illustrates the practicability of the novel approach in terms of preoperative time management, since its employment could save up to nine minutes per vertebra.

Moreover, we also used a spherical template instead of the cubic shaped one, to set up a graph and applied it to vertebra segmentation. As shown in Figure 19 and 20 on the left side, a vertebra can roughly be segmented this way if the density of the rays/sampled nodes and the delta value are set to very large values. However, as soon as these values are smaller, the graph cut prefers a more spherical/elliptical segmentation result, as shown in the middle images of Figure 19 and Figure 20. In the extreme case where the delta value is set to zero, the graph cut has to come back with a perfect sphere and the only variation is the size of the sphere which depends on the gray values (rightmost images of Figure 19 and 20). Hence, the above illustrates the superiority of cubic-shaped templates when it comes to vertebra segmentation.

Furthermore, we also compared the Cube-Cut masks with GrowCut results and found that Cube-Cut clearly outperformed GrowCut in this setup, regardless of the similar DSC values. Besides others, GrowCut regularly did not recognize the vertebral body's outer boundaries in the pedicle-regions, which led to false boundary detections as shown in Figure 21. This strongly illustrates the convenience of the alterable smoothness-term. Furthermore, manual adjustments of the results obtained with Cube-Cut, which at this stage would still be necessary in a clinical context, would take less user effort, since the cubic characteristics of a vertebral body are already incorporated into the calculations. A re-initialization of GrowCut – in case of an unsatisfying segmentation result – can be very time-consuming, because strokes have to be drawn all over again in several 2D slices. The initialization of GrowCut on three 2D slices took an experienced user around one minute and the run time of GrowCut – after the initialization – was around 1-3 minutes. The re-initialization of Cube-Cut however, usually only required the replacement of the one seed.

Nevertheless, visual evaluations of Cube-Cut's segmentation results indicate that the algorithm frequently segments the same specific areas of a vertebral body inaccurately. Recognizing the rectangle shape of a vertebral body, on sagittal slices, these areas could be referred to as the vertebral body's "vertices" (Figure 22). Although the present version of Cube-Cut already allows an arbitrary increase of precision in terms of number of rays and points per ray, future versions of the algorithm could overcome this problem by a densification of rays only in the corresponding spaces or by allowing the user to adjust the segmentation result manually.

Regarding the robustness of Cube-Cut in general, we can report that the method only performs satisfactory if the cube's volume is larger than the vertebral body's (Figure 23, right side), since in case the cube is smaller (Figure 23, left) side, the graph does not penetrate the background which results in an s-t-cut that lies somewhere inside the vertebral body. Thus, a cube larger than the vertebra is desirable. Right now, we estimate the volume of the largest vertebral body in a data set visually and choose the cube's volume accordingly. Future versions of Cube-Cut could provide the user with the possibility of defining the size of the cube interactively, e.g. by drawing a stroke through the vertebra at its largest measurement. The length of the stroke would then be used to define the cube's size automatically, with an additional safety margin.

Furthermore, as already mentioned above, the seed has to be placed roughly around the center region of the vertebral body in order to obtain satisfactory segmentation results. Nevertheless, the algorithm proved itself relatively stable against small deviations and moreover, the replacement of the seed only takes a second which is one of the distinguishing advantages of Cube-Cut.

## Acknowledgements

We want to acknowledge the physicians Thomas Dukatz and Dr. med. Malgorzata Kolodziej from the neurosurgery department of the university hospital in Marburg, Germany, for performing the manual slice-by-slice segmentations of the vertebrae and thus providing the ground truth for the evaluation. Moreover, the authors would like to thank Fraunhofer MeVis in Bremen, Germany, for providing an academic license for MeVisLab. Furthermore, the authors want to thank Bernd Egger for drawing the image on the left side of Figure 2.

| No. | volume of vertebrae (mm³) | | number of voxels | | DSC (%) |
|---|---|---|---|---|---|
| | manual | automatic | manual | automatic | |
| 1 | 23860.6 | 26314.3 | 2927 | 3228 | 86.69 |
| 2 | 27423 | 27431.1 | 3364 | 3365 | 84.17 |
| 3 | 33830.4 | 28776.2 | 4150 | 3530 | 82.06 |
| 4 | 27121.4 | 23901 | 3327 | 2932 | 82.57 |
| 5 | 22165 | 17795.4 | 2719 | 2138 | 71.64 |
| 6 | 15423 | 16638 | 1892 | 2041 | 84.16 |
| 7 | 42658.9 | 33194.5 | 5233 | 4072 | 82.85 |
| 8 | 42715.9 | 35216.2 | 5240 | 4320 | 85.54 |
| 9 | 39903.5 | 29909.3 | 4895 | 3669 | 80.71 |
| 10 | 30594.1 | 18105.4 | 3753 | 2221 | 72.95 |
| min | 15.42 | 16.64 | 1892 | 2041 | 71.64 |
| max | 33.83 | 28.78 | 5240 | 4320 | 86.69 |
| $\mu \pm \sigma$ | 24.97 ± 6.15 | 23.48 ± 5.12 | 3750 | 3152 | 81.33 ± 5.07 |

**Table 1.** Direct comparison of manual slice-by-slice and Cube-Cut segmentation results for ten vertebrae via the Dice Similarity Coefficient (DSC).

| Case | 1 | 2 | 3 | 4 | 5 | 6 | 7 | 8 | 9 | 10 |
|---|---|---|---|---|---|---|---|---|---|---|
| DSC (%) | 78.55 | 81.34 | 83.90 | 71.33 | 71.34 | 70.65 | 88.58 | 91.95 | 85.13 | 83.30 |

**Table 2.** Direct comparison of manual slice-by-slice and GrowCut segmentation results for ten vertebrae via the Dice Similarity Coefficient (note: the cases 1-10 correspond to Table 1).

# Figure Legends

**Figure 1.** T2-weighted MR image showing a degenerative spondylolisthesis (red arrow) and a lumbar stenosis (green arrow).

**Figure 2.** Vertebral anatomy: (a) illustrates the anatomy of a vertebra from a coronal view (adopted from [5]), (b) shows a sagittal T2-weighted MRI slice. The green arrow in the enlargement points to an area inside the vertebral body, whereas the red arrow points to the cortical bone, the outer boundary.

**Figure 3.** Object/background transition regions (red arrows). (a) shows a homogenous object/background transition. In (b), the spinal canal (CSF) makes up parts of the vertebral body's outer boundaries.

**Figure 4.** Illustration of voxel labeling for the foreground ($L_s$) and the background ($L_t$).

**Figure 5.** Profile of two cube faces intersected by three rays (a) and a cubic voxel subset (b).

**Figure 6.** Illustration of the different kinds of edges. (a) i-links: z-edges (black), xy-edges (blue). (b) o-links: s-links(green), t-links(red). (c) whole graph.

**Figure 7.** Illustration of the penalty effect. (a) shows a network without i-links. (b) shows a network with an i-link. The red line depicts a minimal cut.

**Figure 8.** Illustration of the z-edges principle. (a) shows a ray without z-edges: The minimal s-t-cut (red) cuts the ray twice with a capacity of 0. (b) shows the same ray with z-edges. The ray is only cut once. The capacity of the minimal s-t-cut is $\infty + 5$. (c) shows z-edges, embedded into an MR image.

**Figure 9.** Adverse effects on segmentation results (2-dimensional view). (a) shows an overrun in the upper part due to a violation of condition (3). (b) shows a segmentation result affected by an outlier which causes a violation of condition (4). The cut happens too close to the seed point (not shown) in the middle of the vertebra because there is a light area similar to the spinal canal.

**Figure 10.** Effect of the coefficient *w*. In (b), *w* is applied on the s-weights in (a): The cut, with a capacity of $\infty + 2.5$, now happens closer to the seed point. Note that the same cut in (b) would have cost $\infty + 10$ whereas the cut depicted in (b) has a capacity of only $\infty + 5$.

**Figure 11.** Courses of *w(i,11)* (green) and *w(i,15)* (red). The upper part illustrates that *w(i,k)* reflects the position of the voxel $p_i$ on a ray consisting of *k* uniformly distributed voxels. Note that that w is only partially defined for the natural numbers but that Cube-Cut never calls w with an argument in the undefined scope.

**Figure 12.** A feasible surface and intersecting rays (transformed in x-direction for a better visibility). The green node depicts an outlier as it would violate the smoothness constraint $\Delta_x=1$ if classed with the surface voxels.

**Figure 13.** Illustration of the xy-edges principle. (a) shows a minimal cut (thick lines) and the two possible continuations (dashed lines) within the boundaries of a smoothness constraint $\Delta=1$. All other cuts would have a capacity greater than $7 \cdot \infty$. (b) shows the only possible continuation within the boundaries of a $\Delta$-value of 0, where the cut has a capacity of $3 \cdot \infty$.

**Figure 14.** Topology of xy-edges for $\Delta=0$ (a) and $\Delta=1$ (b).

**Figure 15.** Segmentation result for $\Delta=0$ (left) and $\Delta=2$ (right).

**Figure 16.** 3D segmentation result (left and middle image) and 2D segmentation result with the user-defined seed point in blue (rightmost image).

**Figure 17.** Superimposition of a manual segmentation result and a Cube-Cut segmentation result.

**Figure 18.** Typical user initialization of GrowCut for this study. The Editor module is used to mark parts of the vertebra (green) and the background (yellow) in an axial, sagittal and coronal plane.

**Figure 19.** Vertebra segmentation results (red) for a graph that has been constructed with a spherical template from a user-defined seed point (blue). The left image shows the segmentation result when the density of the rays/sampled nodes and the delta value are set to very large values. When these values are smaller the graph cut prefers a more spherical/elliptical segmentation result (middle). The rightmost image shows the extreme case where the delta value was set to zero. There the graph cut has to come back with a perfect sphere and the only variation is the size of the sphere which depends on the gray values.

**Figure 20.** Corresponding 3D results of Figure 19, where a graph has been constructed with a spherical template for vertebra segmentation. The left image shows the 3D segmentation result (yellow) when the density of the rays/sampled nodes and the delta value are set to very large values. When these values are smaller the graph cut prefers a more spherical/elliptical segmentation result (middle). The rightmost image shows the extreme case where the delta value was set to zero, which resulted into a perfect sphere.

**Figure 21.** Sagittal 2D-view on Cube-Cut segmentation result (left, red), GrowCut segmentation result (center, white) and reference image (right). The GrowCut algorithm detects false boundaries in the pedicles-region.

**Figure 22.** "Vertices" of the vertebral body's outer boundaries were not detected accurately (green circles). The upper image shows a 3D segmentation result, the lower images show 2D overlaps of manual (red) and automatic (white) segmentation results.

**Figure 23.** The size of the cube (red) in the left image is too small to segment the vertebra, because a graph that is constructed inside this cube does not cover the border of the vertebra and the s-t-cut will lie inside the vertebra. In contrast, the size of the cube in the right image is sufficient, because a graph that is constructed inside this cube will also cover the vertebra's border and thus is able to segment it.

**Figures**

Figure 1

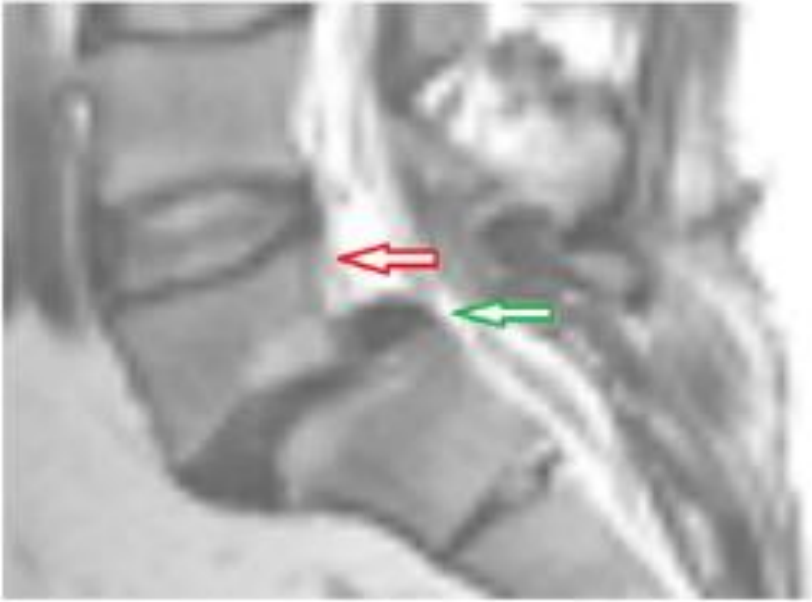

Figure 2

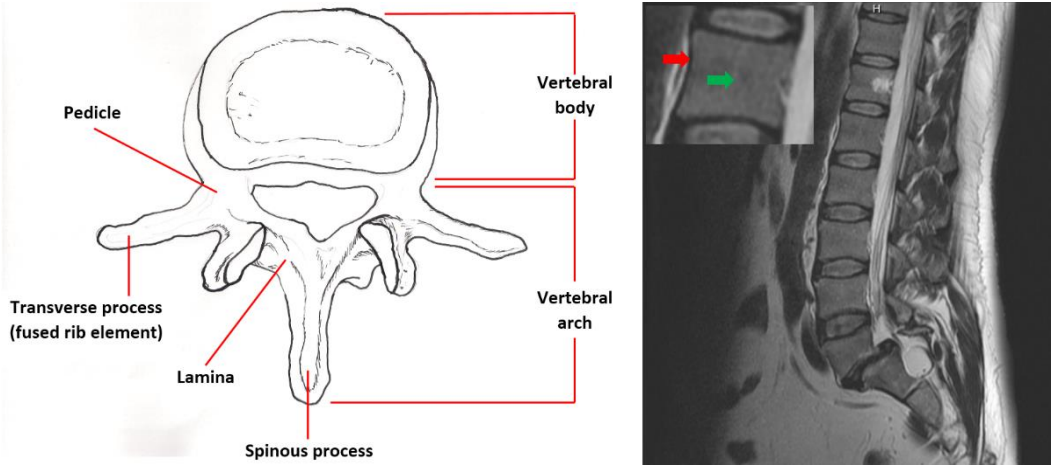

Figure 3

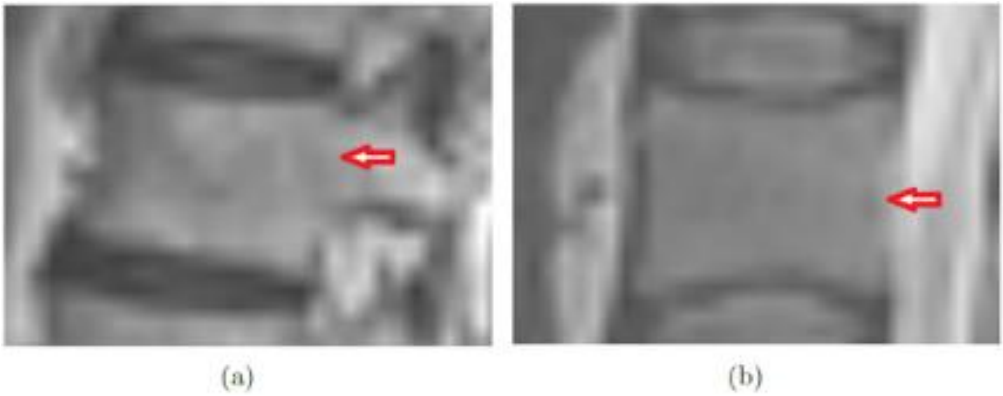

Figure 4

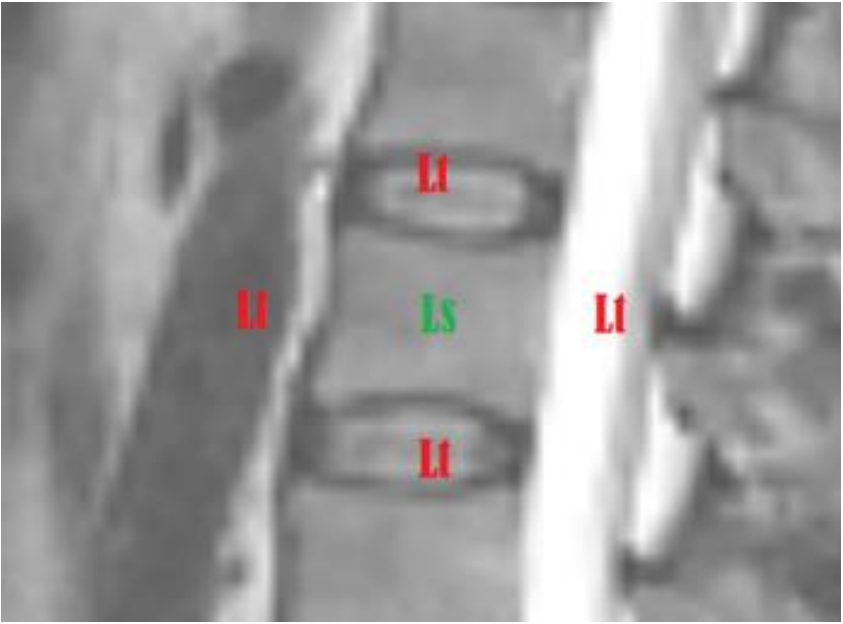

Figure 5

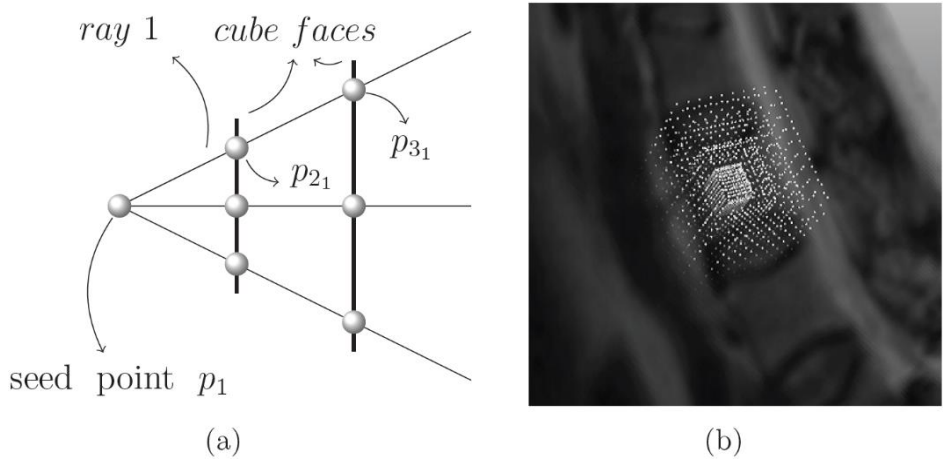

Figure 6

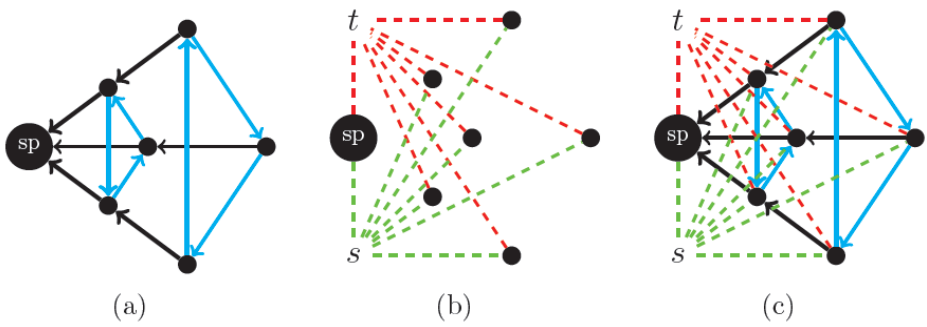

Figure 7

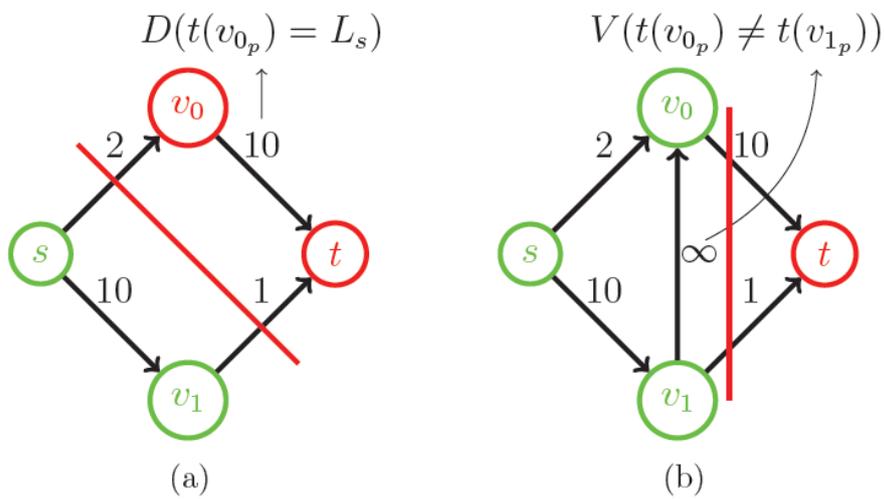

Figure 8

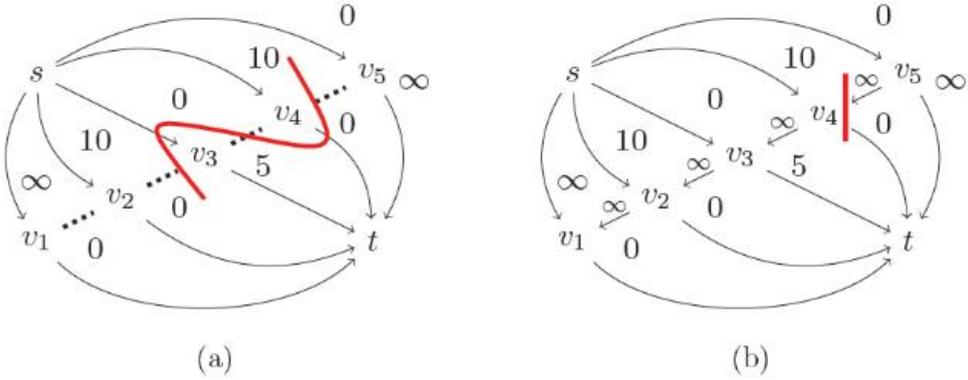

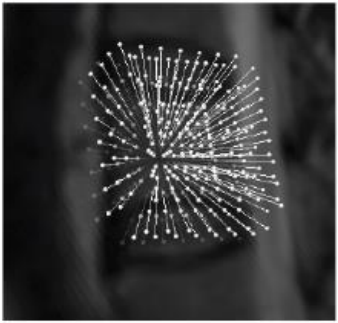

Figure 9

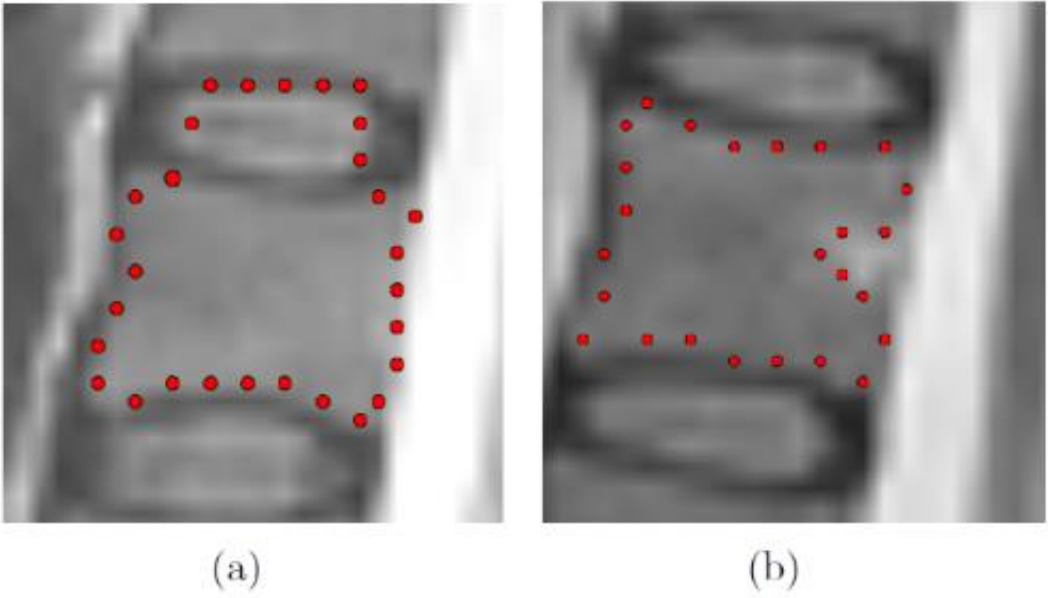

Figure 10

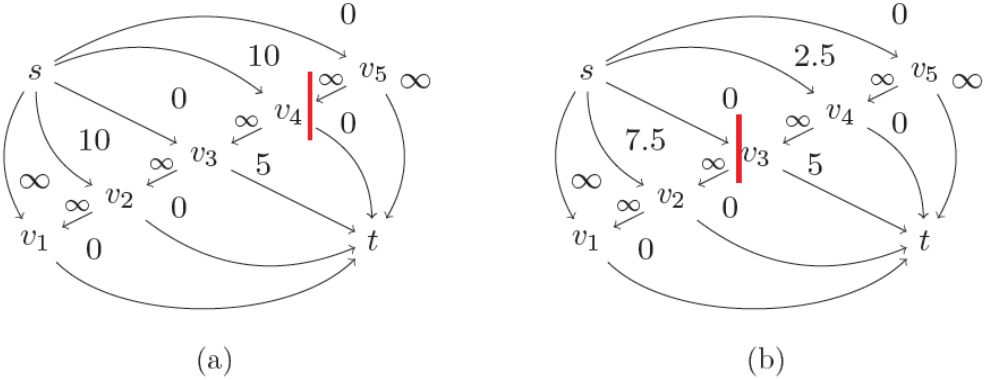

Figure 11

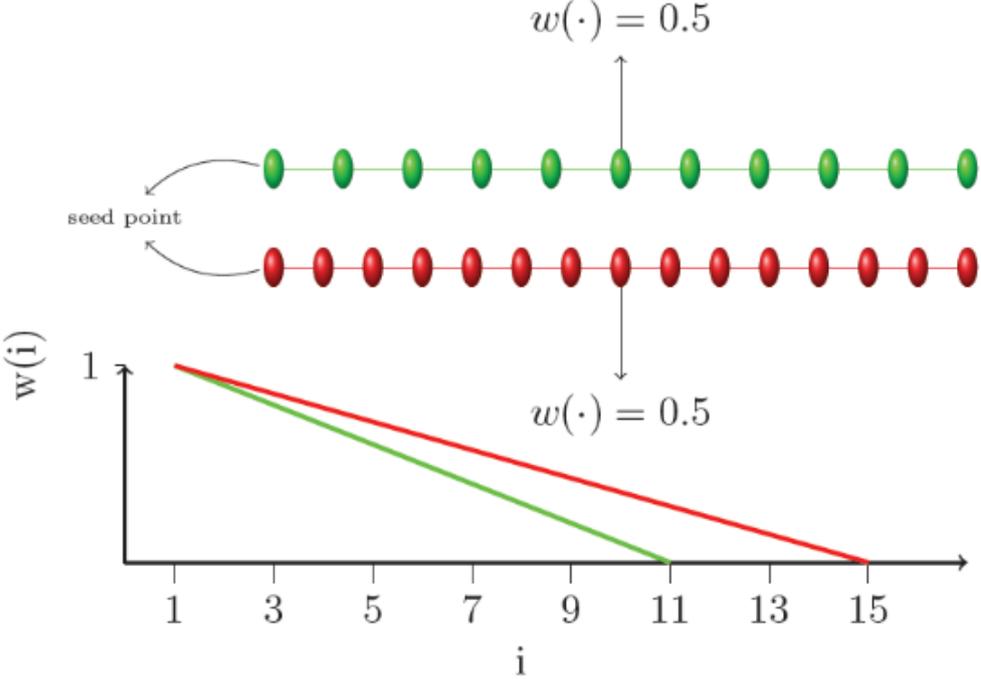

Figure 12

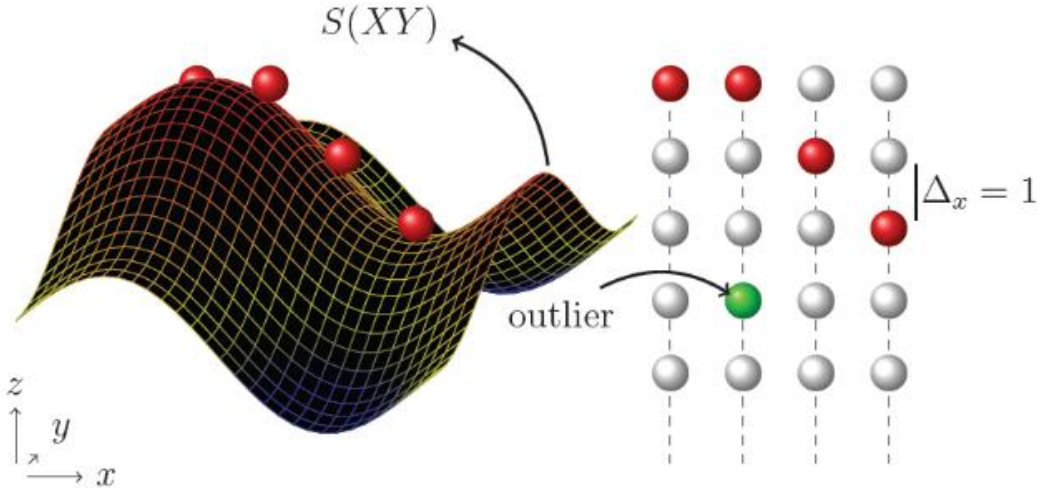

Figure 13

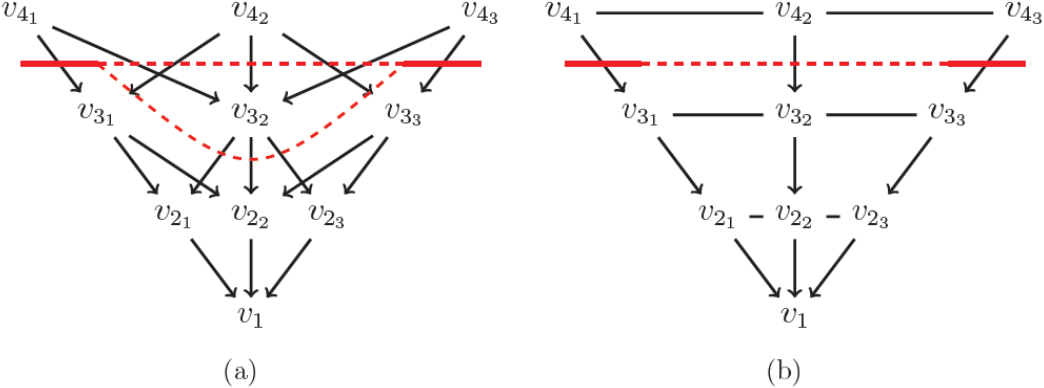

Figure 14

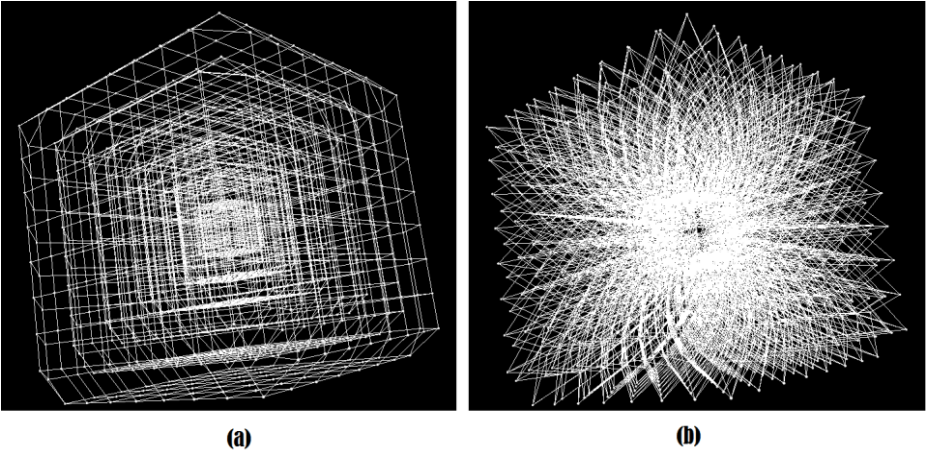

Figure 15

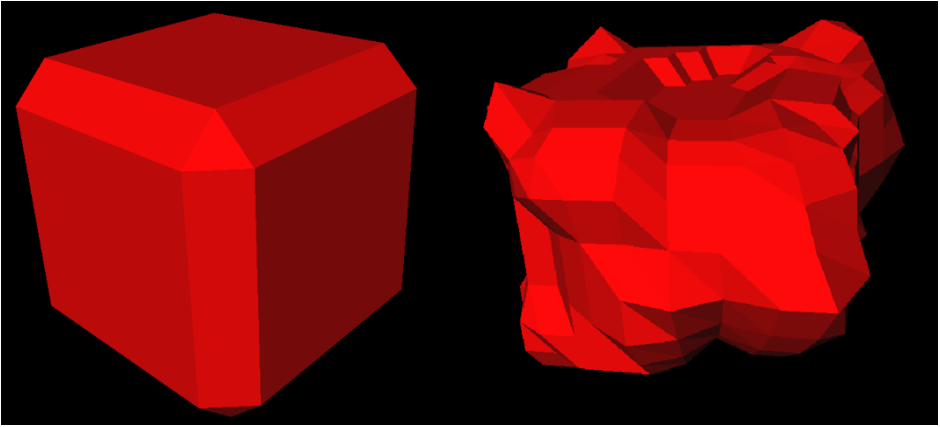

Figure 16

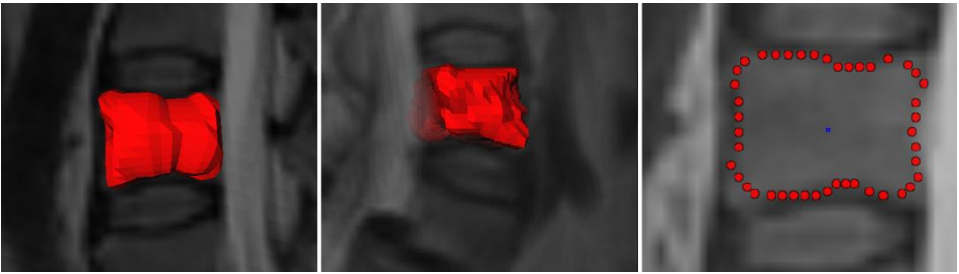

Figure 17

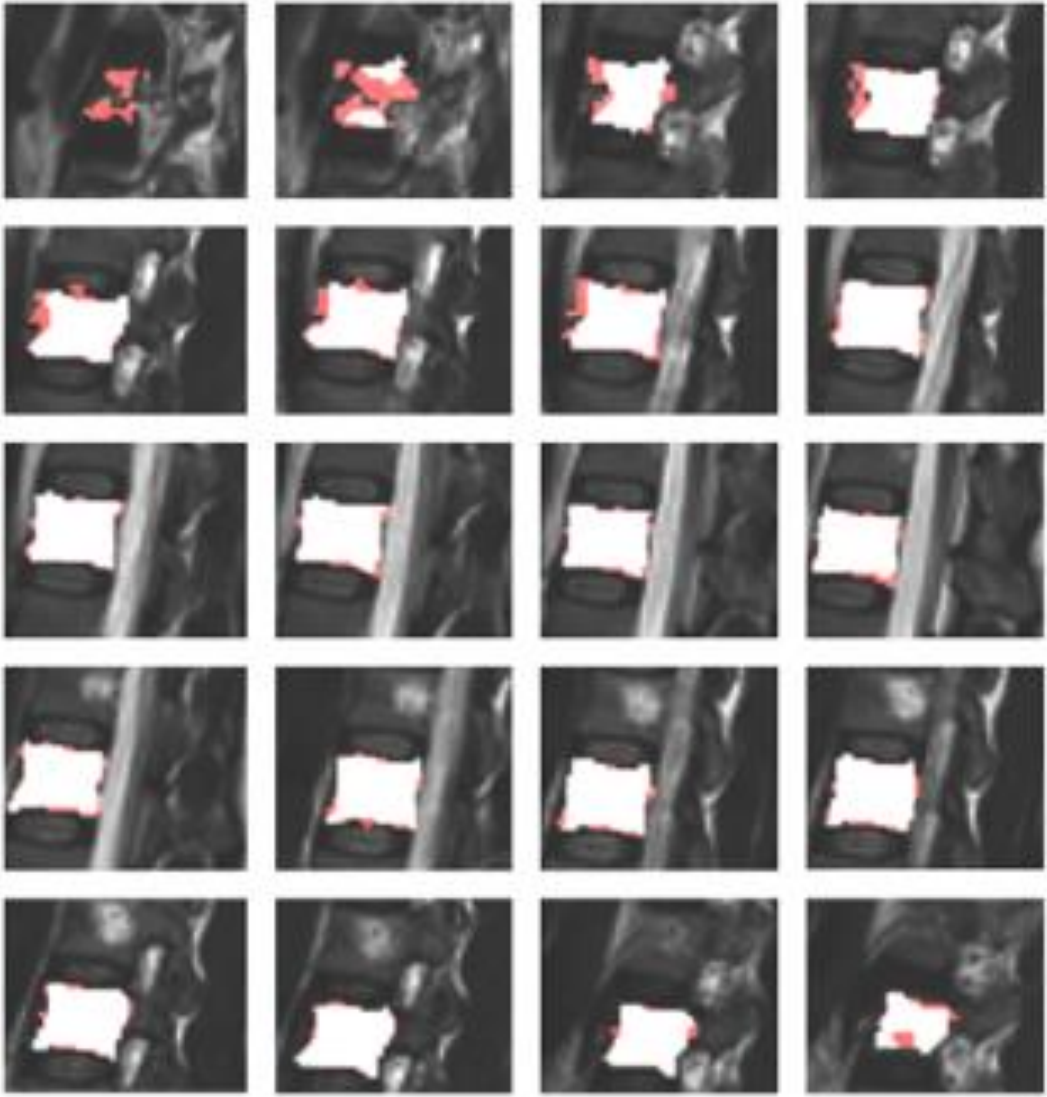

Figure 18

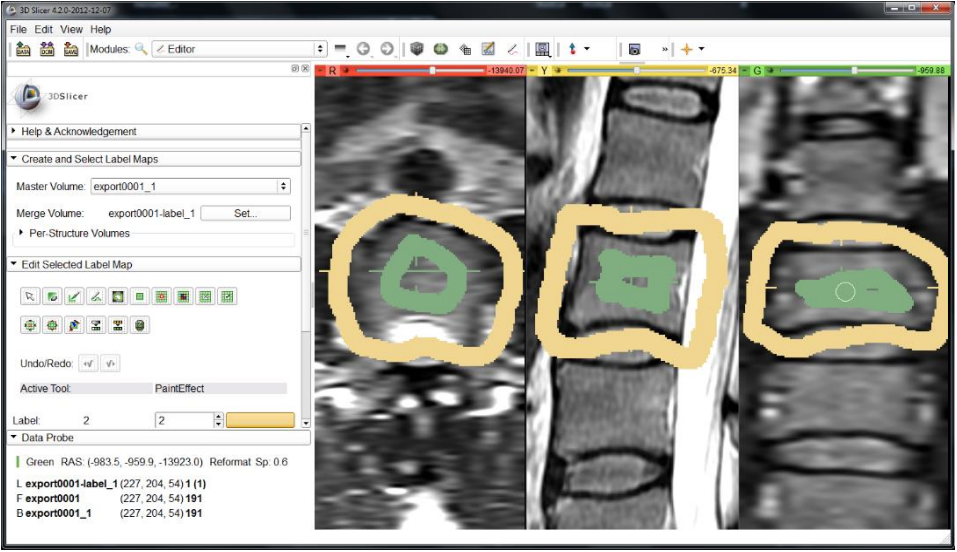

Figure 19

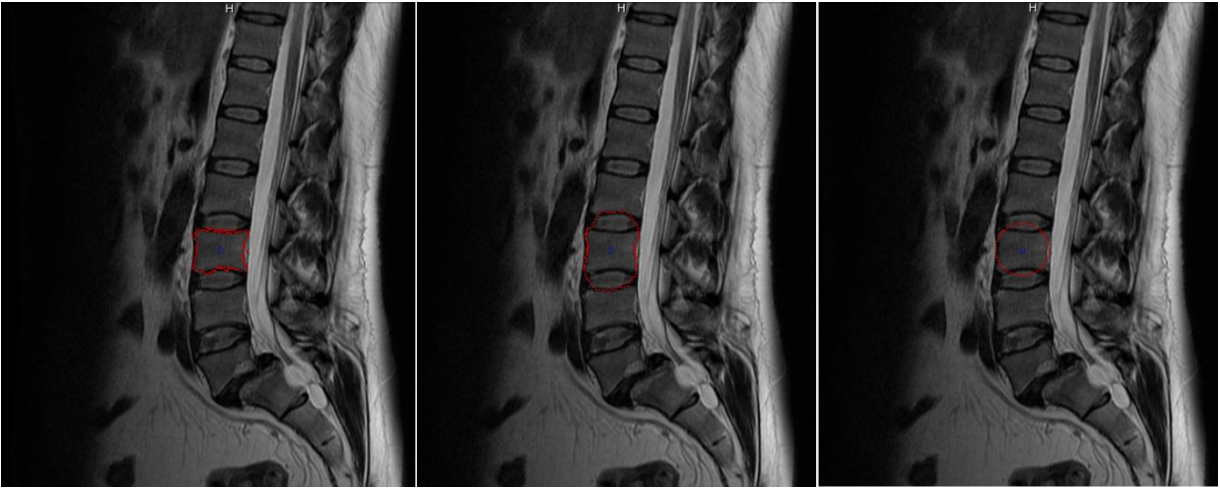

Figure 20

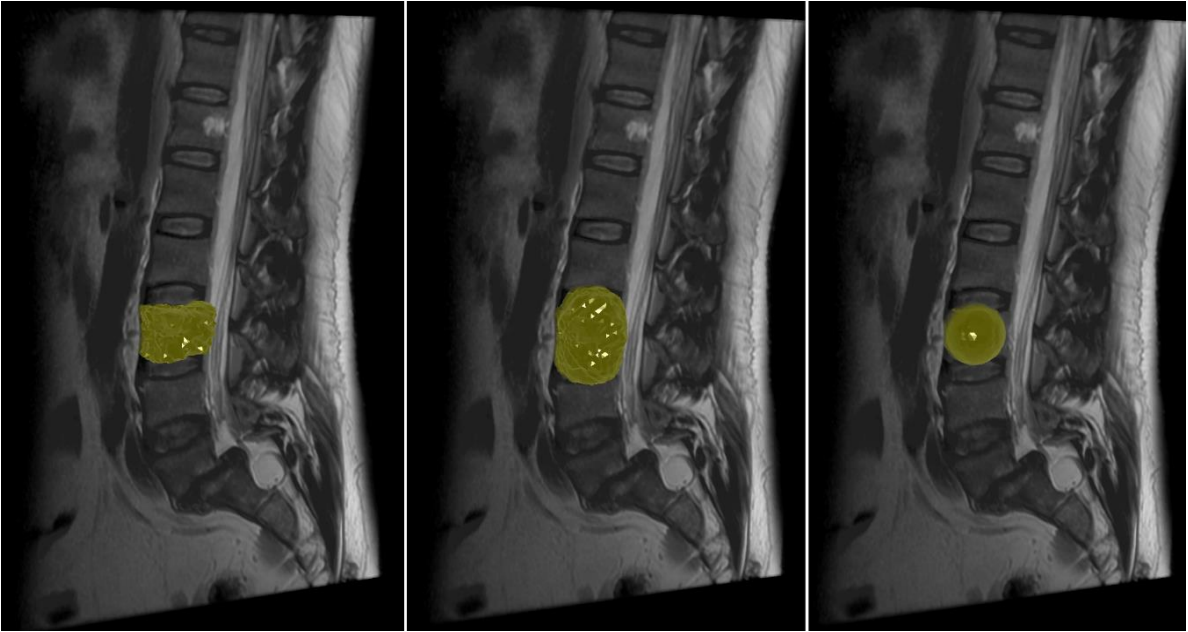

Figure 21

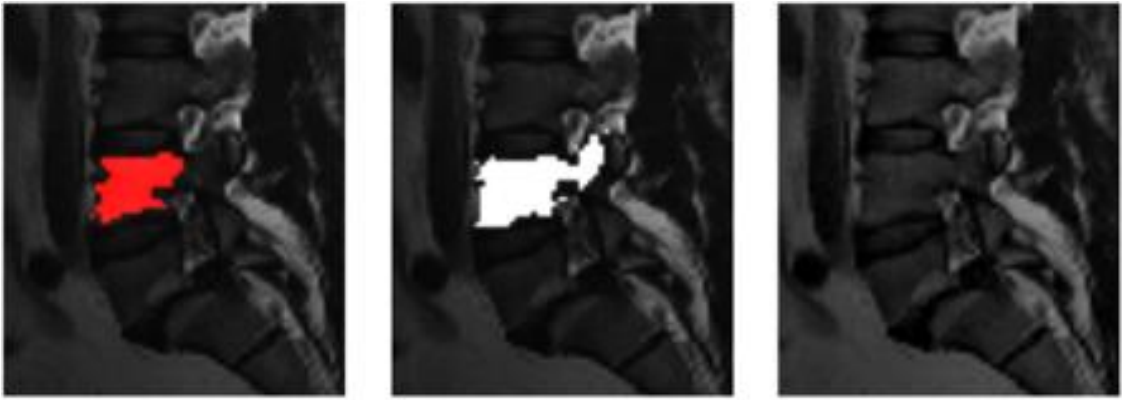

Figure 22

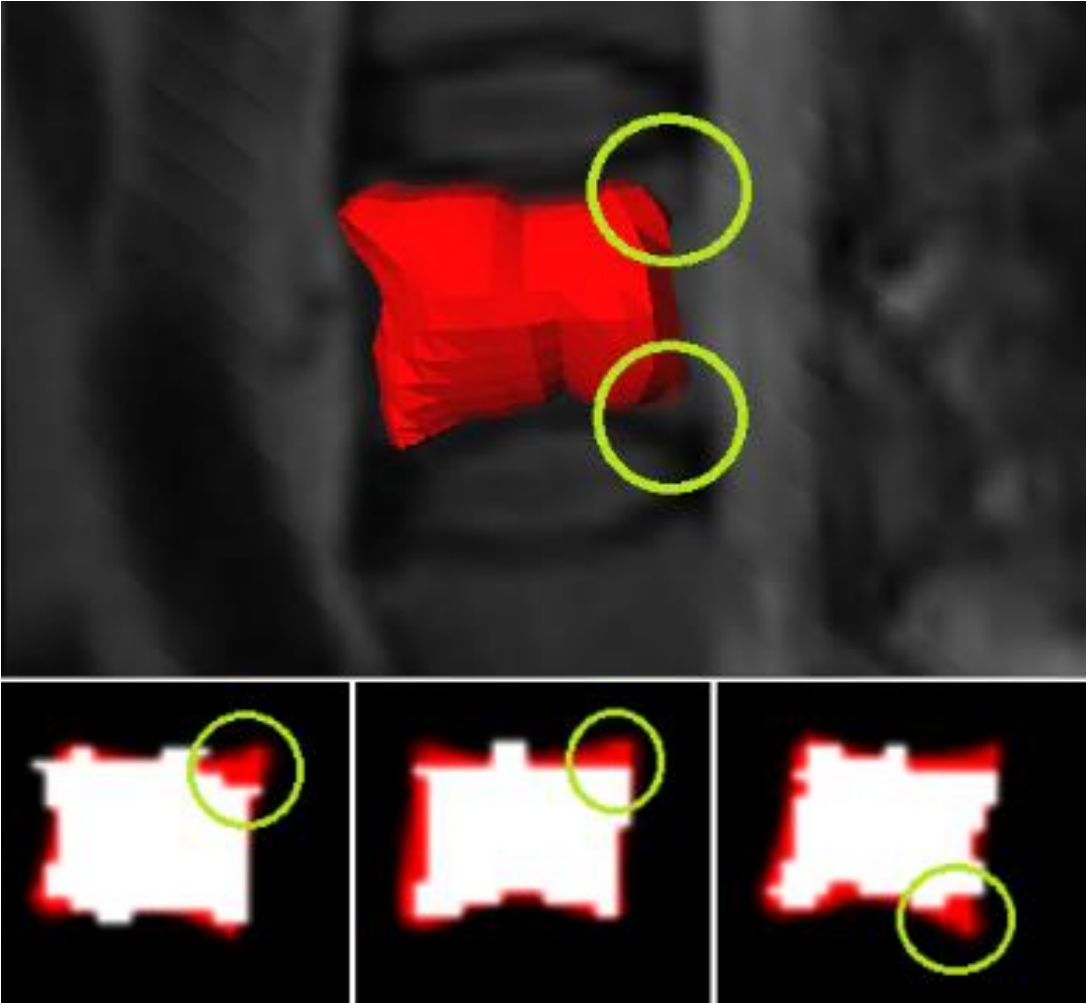

Figure 23

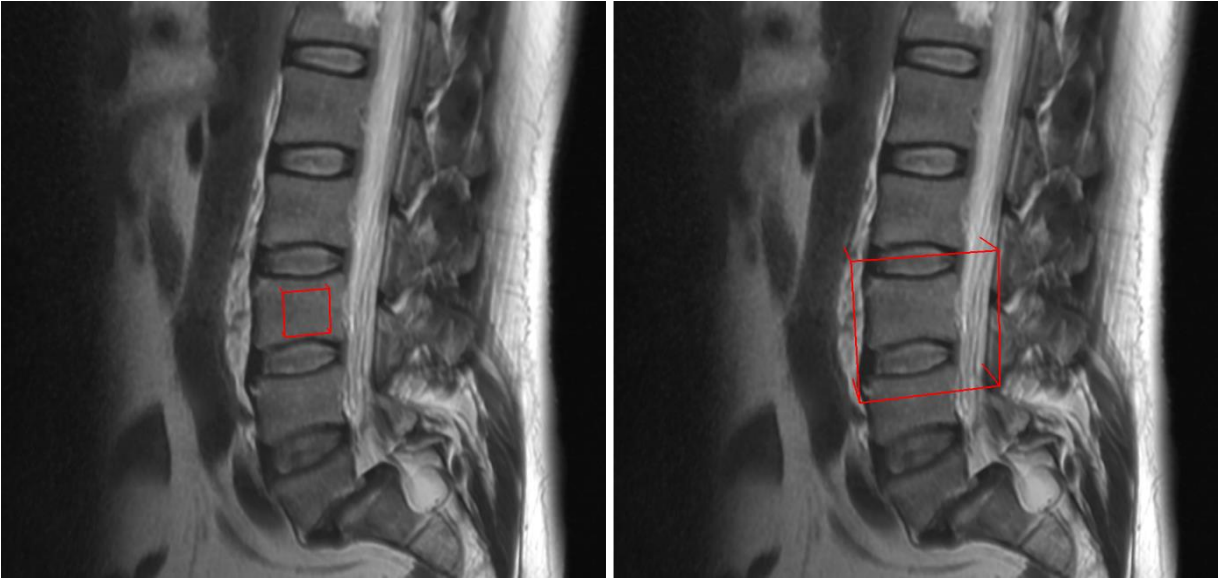